\documentclass{acmsmall} 

\usepackage[ruled]{algorithm2e}

\usepackage{times}
\usepackage{url}
\usepackage{latexsym}
\usepackage{multirow}
\usepackage{graphicx}
\usepackage{times}
\usepackage{latexsym}
\usepackage{amsmath}
\usepackage{amssymb}
\usepackage{paralist}
\usepackage{url}
\usepackage{paralist}
\usepackage{booktabs}
\usepackage[caption=false]{subfig}
\usepackage{epstopdf}
\usepackage[numbers]{natbib}

\usepackage{tikz}

\newcommand\idea[1]{\textcolor{olive}{}} 

\newcommand\work{paper }

\usepackage{color}

\begin{document}

\markboth{M. Lukasik et al.}{Using Gaussian Processes for Rumour Stance Classification in Social Media}

\title{Using Gaussian Processes for Rumour Stance Classification in Social Media}
\author{MICHAL LUKASIK
\affil{University of Sheffield}
KALINA BONTCHEVA
\affil{University of Sheffield}
TREVOR COHN
\affil{University of Melbourne}
ARKAITZ ZUBIAGA
\affil{University of Warwick}
MARIA LIAKATA
\affil{University of Warwick}
ROB PROCTER
\affil{University of Warwick}
}

\begin{abstract}
Social media tend to be rife with rumours while new reports are released piecemeal during breaking news. Interestingly, one can mine multiple reactions expressed by social media users in those situations, exploring their stance towards rumours, ultimately enabling the flagging of highly disputed rumours as being potentially false. In this work, we set out to develop an automated, supervised classifier that uses multi-task learning to classify the stance expressed in each individual tweet in a rumourous conversation as either supporting, denying or questioning the rumour. Using a classifier based on Gaussian Processes, and exploring its effectiveness on two datasets with very different characteristics and varying distributions of stances, we show that our approach consistently outperforms competitive baseline classifiers. Our classifier is especially effective in estimating the distribution of different types of stance associated with a given rumour, which we set forth as a desired characteristic for a rumour-tracking system that will warn both ordinary users of Twitter and professional news practitioners when a rumour is being rebutted.
\end{abstract}



%

\maketitle

\section{Introduction}

There is an increasing need to interpret and act upon rumours spreading quickly through social media during breaking news, where new reports are released piecemeal and often have an unverified status at the time of posting. Previous research has posited the damage that the diffusion of false rumours can cause in society, and that corrections issued by news organisations or state agencies such as the police may not necessarily achieve the desired effect sufficiently quickly \cite{lewandowsky2012misinformation,procter2013readingb}. Being able to determine the accuracy of reports is therefore crucial in these scenarios. However, the veracity of rumours in circulation is usually hard to establish \cite{allport1947}, since as many views and testimonies as possible need to be assembled and examined in order to reach a final judgement. Examples of rumours that were later disproven, after being widely circulated, include a 2010 earthquake in Chile, where rumours of a volcano eruption and a tsunami warning in Valparaiso spawned on Twitter \cite{Mendoza-rumours-2010}. Another example is the England riots in 2011, where false rumours claimed that rioters were going to attack Birmingham's Children's Hospital and that animals had escaped from London Zoo \cite{procter2013readinga}.

Previous work by ourselves and others has argued that looking at how users in social media orient to rumours is a crucial first step towards making an informed judgement on the veracity of a rumourous report \cite{zubiaga2016analysing,tolmie2015microblog,Mendoza-rumours-2010}. For example, in the case of the riots in England in August 2011, Procter et al. manually analysed the stance expressed by users in social media towards rumours \cite{procter2013readinga}. Each tweet discussing a rumour was manually categorised as supporting, denying or questioning it. It is obvious that manual methods have their disadvantages in that they do not scale well; the ability to perform stance categorisation of tweets in an automated way would be of great use in tracking rumours, flagging those that are largely denied or questioned as being more likely to be false. 

Determining the stance of social media posts automatically has been attracting increasing interest in the scientific community in recent years, as this is a useful first step towards more in-depth rumour analysis:
\begin{itemize}
\item It can help detect rumours and flag them as such more quickly \cite{zhao2015www}.
\item It is useful for tracking public opinion about rumours and hence for monitoring their wider effect on society.
\item Aggregate stance information and dynamics over time can be leveraged for rumour veracity classification \cite{derczynski2015pheme,Liu2015_rumourdebunking_cikm}.
\end{itemize} 
Work on automatic rumour stance classification, however, is still in its infancy, with some methods ignoring temporal ordering and rumour identities (e.g. \cite{Qazvinian:2011:RIM:2145432.2145602}), while others being rule-based and thus with unclear generalisability to new rumours \cite{zhao2015www}.

Our work advances the state-of-the-art in tweet-level stance classification through multi-task learning and Gaussian Processes. This article substantially extends our earlier short paper \cite{Lukasik15_qazvinian}, fistly by using a second dataset, which enables us to test the generalisability of our results. Secondly, a comparison against additional baseline classifiers and recent state-of-the-art approaches has been added to the experimental section. Lastly, we carried out a more thorough analysis of the results, including now per-class performance scores, which furthers our understanding of rumour stance classification.

\begin{table}
\tbl{Tweets pertaining to a rumour about hospital being attacked during 2011 England Riots.\label{tab:tweets_hospital}}{
\begin{tabular}{p{0.75\linewidth}l}
\toprule text & position\\
\midrule
Birmingham Children's hospital has been attacked. F***ing morons. \#UKRiots & support \\
\midrule
Girlfriend has just called her ward in Birmingham Children's
Hospital \& there's no sign  of any trouble \#Birminghamriots & deny \\
\midrule
Birmingham children's hospital guarded by police? Really? Who would target a childrens hospital \#disgusting \#Birminghamriots & question \\
\bottomrule
\end{tabular}}
\end{table}

In comparison to the state-of-the-art, our approach is novel in several crucial aspects:
\begin{enumerate}
  \item We perform stance classification on \emph{unseen rumours}, given a training set of already annotated rumours on different topics and from different time periods.
  \item The \emph{temporal ordering of tweets} on a given rumour is respected, both during training and stance classification.
  \item \emph{Generalisability to new datasets} is a core aspect of our methodology, which is built on the premise that patterns of stance should exhibit similar characteristics across different rumours.
\end{enumerate}

Based on the assumption of a common underlying linguistic signal in rumours on different topics, we build a transfer learning system based on Gaussian Processes, that can classify stance in newly emerging rumours. The paper reports results on two different rumour datasets and explores two different experimental settings -- without any training data and with very limited training data. We refer to these as:
\begin{itemize}
\item \emph{Leave One Out}: all tweets pertaining to a target rumour are only used for testing, i.e. method performance on a completely unseen rumour is reported; 
\item \emph{Leave Part Out}: the first few tweets of a target rumour (as annotated by journalists) and added to the training set of the Gaussian Process classifier, together with tweets pertaining to older rumours. The rest of the tweets on the target rumour are used for evaluation. 
\end{itemize}

Our results demonstrate that Gaussian Process-based, multi-task learning leads to significantly improved performance over state-of-the-art methods and competitive baselines, as demonstrated on two very different datasets. The classifier relying on Gaussian Processes performs particularly well over the rest of the baseline classifiers in the Leave Part Out setting, proving that it does particularly well in determining the distribution of supporting, denying and questioning tweets associated with a rumour. Estimating the distribution of stances is the key aspect for which our classifier performs especially well compared to the baseline classifiers.

\section{Related Work}
\label{sec:rel_work}

This section provides a more in-depth motivation of the rumour stance detection task and an overview of the state-of-the-art methods and their limitations. First, however, let us start by introducing the formal definition of a rumour.

\subsection{Rumour Definition}

There have been multiple attempts at defining rumours in the literature. Most of them are complementary to one another, with slight variations depending on the context of their analyses. The core concept that most researchers agree on matches the definition that major dictionaries provide, such as the Oxford English Dictionary\footnote{http://www.oxforddictionaries.com/definition/english/rumour} defining a rumour as \emph{``a currently circulating story or report of uncertain or doubtful truth''}. For instance, DiFonzo and Bordia \cite{difonzo2007rumor} defined rumours as ``unverified and instrumentally relevant information statements in circulation.''

Researchers have long looked at the properties of rumours to understand their diffusion patterns and to distinguish them from other kinds of information that people habitually share \cite{donovan2007idle}. Allport and Postman \cite{allport1947} claimed that rumours spread due to two factors: people want to find meaning in things and, when faced with ambiguity, people try to find meaning by telling stories. The latter factor also explains why rumours tend to change in time by becoming shorter, sharper and more coherent. This is the case, it is argued, because in this way rumours explain things more clearly. On the other hand, Rosnow \cite{rosnow1991} claimed that there are four important factors for rumour transmission. Rumours must be outcome-relevant to the listener, must increase personal anxiety, be somewhat credible and be uncertain. Furthermore, Shibutani \cite{shibutani1969} defined rumours to be \emph{“a recurrent form of communication through which men [sic] caught together in an ambiguous situation attempt to construct a meaningful interpretation of it by pooling their intellectual resources. It might be regarded as a form of collective problem-solving”}. 

In contrast with these three theories, Guerin and Miyazaki \cite{guerin2006} state that a rumour is a form of relationship-enhancing talk. Building on their previous work, they recall that many ways of talking serve the purpose of forming and maintaining social relationships. Rumours, they say, can be explained by such means. 

In our work, we adhere to the widely accepted fact that rumours are unverified pieces of information. More specifically, following \cite{zubiaga2016analysing}, we regard a rumour in the context of breaking news, as a \emph{``circulating story of questionable veracity, which is apparently credible but hard to verify, and produces sufficient skepticism and/or anxiety so as to motivate finding out the actual truth''}.

\subsection{Descriptive Analysis of Rumours in Social Media}

One particularly influential piece of work in the field of rumour analysis in social media is that by Mendoza et al. \cite{Mendoza-rumours-2010}. 
By manually analysing the data from the earthquake in Chile in 2010, the authors selected 7 confirmed truths and 7 false rumours, each consisting of close to 1000 tweets or more. The veracity value of the selected stories was corroborated by using reliable sources. 
Each tweet from each of the news items was manually classified into one of the following classes: affirmation, denial, questioning, unknown or unrelated. In this way, each tweet was classified according to the position it showed towards the topic it was about. The study showed that a much higher percentage of tweets about false rumours are shown to deny the respective rumours (approximately 50\%). This is in contrast to rumours later proven to be true, where only 0.3\% of tweets were denials. Based on this, authors claimed that rumours can be detected using aggregate analysis of the stance expressed in tweets.

Recent research put together in a special issue on rumours and social media \cite{papadopoulos2016overview} also shows the increasing interest of the scientific community in the topic. \cite{webb2016digital} proposed an agenda for research that establishes an interdisciplinary methodology to explore in full the propagation and regulation of unverified content on social media. \cite{middleton2016geoparsing} described an approach for geoparsing social media posts in real-time, which can be of help to determine the veracity of rumours by tracking down the poster's location. The contribution of \cite{hamdi2016tison} to rumour resolution is to build an automated system that rates the level of trust of users in social media, hence enabling to get rid of users with low reputation. Complementary to these approaches, our objective is to determine the stance of tweets towards a rumour, which can then be aggregated to establish an overall veracity score for the rumour.

Another study that shows insightful conclusions with respect to stance towards rumours is that by Procter et al. \cite{procter2013readinga}.
The authors conducted an analysis of a large dataset of tweets related to the riots in the UK, which took place in August 2011. The dataset collected in the riots study is one of the two used in our experiments, and we describe it in more detail in section \ref{sec:data}. After grouping the tweets into topics, where each represents a rumour, they were manually categorised into different classes, namely:
\begin{inparaenum}
\item media reports, which are tweets sent by mainstream media accounts or journalists connected to media,
\item pictures, being tweets uploading a link to images,
\item rumours, being tweets claiming or counter claiming something without giving any source,
\item reactions, consisting of tweets being responses of users to the riots phenomenon or specific event related to the riots.
\end{inparaenum}

Besides categorisation of tweets by type, Procter et al. also manually categorised the accounts posting tweets into different types, such as mainstream media, only on-line media, activists, celebrities, bots,  among others.

What is interesting for the purposes of our work is that the authors observed the following four-step pattern recurrently occurring across the collected rumours:
\begin{inparaenum}
\item a rumour is initiated by someone claiming it may be true,
\item a rumour spreads together with its reformulations,
\item counter claims appear,
\item a consensus emerges about the credibility of the rumour.
\end{inparaenum}

This leads the authors to the conclusion that the process of 'inter-subjective sense making' by Twitter users plays a key role in exposing false rumours. This finding, together with subsequent work by Tolmie et al. into the conversational characteristics of microblogging \cite{tolmie2015microblog} has motivated our research into automating stance classification as a methodology for accelerating this process.

\subsection{Rumour Stance Classification}

Qazvinian et al. \cite{Qazvinian:2011:RIM:2145432.2145602} conducted early work on rumour stance classification. They introduced a system that analyzes a set of tweets associated with a given topic predefined by the user. Their system would then classify each of the tweets as supporting, denying or questioning a tweet. We have adopted this scheme in terms of the different types of stance in the work we report here. However, their work ended up merging denying and questioning tweets for each rumour into a single class, converting it into a 2-way classification problem of supporting vs denying-or-questioning. Instead, we keep those classes separate and, following Procter et al., we conduct a 3-way classification \cite{zubiaga2014d2}. 

Another important characteristic that differentiates Qazvinian et al.'s work from ours is that they looked at support and denial on longstanding rumours, such as the fact that many people conjecture whether Barack Obama is a Muslim or not. By contrast, we look at rumours that emerge in the context of fast-paced, breaking news situations, where new information is released piecemeal, often with statements that employ hedging words such as ``reportedly'' or ``according to sources'' to make it clear that the information is not fully verified at the time of posting. This is a very different scenario from that in Qazvinian et al.'s work as the emergence of rumourous reports can lead to sudden changes in vocabulary, leading to situations that might not have been observed in the training data.

Another aspect that we deal with differently in our work, aiming to make it more realistically applicable to a real world scenario, is that we apply the method to each rumour separately. Ultimately, our goal is to classify new, emerging rumours, which can differ from what the classifier has observed in the training set. Previous work ignored this separation of rumours, by pooling together tweets from all the rumours in their collections, both in training and test data. By contrast, we consider the rumour stance classification problem as a form of transfer learning and seek to classify unseen rumours by training the classifier from previously labelled rumours. We argue that this makes a more realistic classification scenario towards implementing a real-world rumour-tracking system.

Following a short gap, there has been a burst of renewed interest in this task since 2015. For example, Liu et al. \cite{Liu2015_rumourdebunking_cikm} introduce rule-based methods for stance classification, which were shown to outperform the approach by \cite{Qazvinian:2011:RIM:2145432.2145602}. Similarly, \cite{zhao2015www} use regular expressions instead of an automated method for rumour stance classification. Hamidian and Diab \cite{hamidian2016rumor} use Tweet Latent Vectors to assess the ability of performing 2-way classification of the stance of tweets as either supporting or denying a rumour. They study the extent to which a model trained on historical tweets can be used for classifying new tweets on the same rumour. This, however, limits the method's applicability to long-running rumours only.

The work closest to ours in terms of aims is Zeng et al. \cite{zeng2016unconfirmed}, who explored the use of three different classifiers for automated rumour stance classification on unseen rumours. In their case, classifiers were set up on a 2-way classification problem dealing with tweets that support or deny rumours. In the present work, we extend this research by performing 3-way classification that also deals with tweets that question the rumours. Moreover, we adopt the three classifiers used in their work, namely Random Forest, Naive Bayes and Logistic Regression, as baselines in our work.

Lastly, researchers \cite{zhao2015www,Ma2015_detectrumours_cikm} have focused on the related task of detecting rumours in social media. While a rumour detection system could well be the step that is applied prior to our stance classification system, here we assume that rumours have already been identified to focus on the subsequent step of determining stances.

\section{Problem Definition: Tweet Level Rumour Stance Classification}

\subsection{Definition of the Task}
\label{ssec:task}

Individual tweets may discuss the same rumour in different ways, where each user expresses their own stance towards the rumour. Within this scenario, we define the tweet level rumour stance classification task as that in which a classifier has to determine the stance of each tweet towards the rumour. More specifically, given the tweet $t_i$ as input, the classifier has to determine which of the set $Y = \{supporting, denying, questioning\}$ applies to the tweet, $y(t_i) \in Y$.

Here we define the task as a supervised classification problem, where the classifier is trained from a labelled set of tweets and is applied to tweets on a new, unseen set of rumours.

\subsection{Problem formulation}
\label{ssec:problem}

Let $R$ be a set of rumours, each of which consists of tweets discussing it, $\forall_{r \in R}$ $T_r$ $= \{t^r_1, \cdots, t^r_{r_n}\}$. $T = \cup_{r \in R} T_r$ is the complete set of tweets from all rumours. Each tweet is classified as supporting, denying or questioning with respect to its rumour: $y(t_i) \in \{s, d, q\}$.

We formulate the problem in two different settings. First, we consider the Leave One Out (LOO) setting, which means that for each rumour $r \in R$, we construct the test set equal to $T_r$ and the training set equal to $T \setminus T_r$. This is the most challenging scenario, where the test set contains an entirely unseen rumour.

The second setting is Leave Part Out (LPO). In this formulation, a very small number of initial tweets from the target rumour is added to the training set $\{t^r_1, \cdots, t^r_{{{r_k}}}\}$.
This scenario becomes applicable typically soon after a rumour breaks out and journalists have started monitoring and analysing the related tweet stream.
The experimental section investigates how the number of initial training tweets influences classification performance on a fixed test set, namely: $\{t^r_{{{r_l}}{}}, \cdots, t^r_{r_n}\}$, $l>k$.

The tweet-level stance classification problem here assumes that tweets from the training set are already labelled with the rumour discussed and the attitude expressed towards that. This information can be acquired either via manual annotation as part of expert analysis, as is the case with our dataset, or automatically, e.g. using pattern-based rumour detection \cite{zhao2015www}. Our method is then used to classify the stance expressed in each new tweet from the test set. 

\subsection{Datasets}
\label{sec:data}
\newcommand\no{$\times$}

\begin{table}
\centering
\tbl{Counts of tweets with supporting, denying or questioning labels in each rumour collection from the England riots dataset.\label{table:Datacounts}}{
\begin{tabular}{lrrr}
\toprule 
Rumour & Supporting & Denying & Questioning \\ \hline
\midrule 
Army bank & 62 & 42 & 73 \\
Children's hospital & 796 & 487 & 132 \\
London Eye & 177 & 295 & 160 \\
McDonald's & 177 & 0 & 13 \\
Miss Selfridge's & 3150 & 0 & 7 \\
Police beat girl & 783 & 4 & 95 \\
London zoo & 616 & 129 & 99 \\
\midrule
Total & 5761 & 957 & 579 \\
\bottomrule 
\end{tabular}}
\end{table}

We evaluate our work on two different datasets, which we describe below. We use two recent datasets from previous work for our study, both of which adapt to our needs. We do not use the dataset by \cite{Qazvinian:2011:RIM:2145432.2145602} given that it uses a different annotation scheme limited to two categories of stances.

The reason why we use the two datasets separately instead of combining them is that they have very different characteristics. Our experiments, instead, enable us to assess the ability of our classifier to deal with these different characteristics.

\subsubsection{England riots dataset}

The first dataset consists of several rumours circulating on Twitter during the England riots in 2011 (see Table~\ref{table:Datacounts}). The dataset was collected by tracking a long set of keywords associated with the event. The dataset was analysed and annotated manually as supporting, questioning, or denying a rumour, by a team of social scientists studying the role of social media during the riots \cite{procter2013readinga}. 

As can be seen from the dataset overview in Table~\ref{table:Datacounts}, different rumours exhibit varying proportions of supporting, denying and questioning tweets, which was also observed in other studies of rumours \cite{Mendoza-rumours-2010,Qazvinian:2011:RIM:2145432.2145602}.
These variations in the number of instances for each class across rumours posits the challenge of properly modelling a rumour stance classifier. The classifier needs to be able to deal with a test set where the distribution of classes can be very different to that observed in the training set.

Thus, we perform 7-fold cross-validation in the experiments, each fold having six rumours in the training set, and the remaining rumour in the test set.

The seven rumours were as follows \cite{procter2013readinga}:
\begin{itemize}
    \item Rioters had attacked London Zoo and released the animals.
    \item Rioters were gathering to attack Birmingham's Children's Hospital.
    \item Rioters had set the London Eye on fire.
    \item Police had beaten a sixteen year old girl.
    \item The Army was being mobilised in London to deal with the rioters.
    \item Rioters had broken into a McDonalds and set about cooking their own food.
    \item A store belonging to the Miss Selfridge retail group had been set on fire in Manchester.
\end{itemize}

\idea{Refer to Figure~\ref{fig:temporal_points}}

\subsubsection{PHEME dataset}

Additionally, we use another rumour dataset associated with five different events, which was collected as part of the PHEME FP7 research project and described in detail in \cite{zubiaga2016analysing,d24_deliverable}. Note that the authors released datasets for nine events, but here we remove non-English datasets, as well as small English datasets each of which includes only 1 rumour, as opposed to the 40+ rumours in each of the datasets that we are using. We summarise the details of the five events we use from this dataset in Table~\ref{table:DatacountsPheme}.

In contrast to the England riots dataset, the PHEME datasets were collected by tracking conversations initiated by rumourous tweets. This was done in two steps. First, we collected tweets that contained a set of keywords associated with a story unfolding in the news. We will be referring to the latter as an event. 
Next, we sampled the most retweeted tweets, on the basis that rumours by definition should be ``a circulation story which produces sufficient skepticism or anxiety''. This allows us to filter  potentially rumourous tweets and collect conversations initiated by those. Conversations were tracked by collecting replies to tweets and, therefore, unlike the England riots, this dataset also comprises replying tweets by definition. This is an important characteristic of the dataset, as one would expect that replies are generally shorter and potentially less descriptive than the source tweets that initiated the conversation. We take this difference into consideration when performing the analysis of our results.

This dataset includes tweets associated with the following five events:

\begin{itemize}
 \item \textbf{Ferguson unrest:} Citizens of Ferguson in Michigan, USA, protested after the fatal shooting of an 18-year-old African American, Michael Brown, by a white police officer on August 9, 2014.
 \item \textbf{Ottawa shooting:} Shootings occurred on Ottawa's Parliament Hill in Canada, resulting in the death of a Canadian soldier on October 22, 2014.
 \item \textbf{Sydney siege:} A gunman held as hostages ten customers and eight employees of a Lindt chocolate caf\'e located at Martin Place in Sydney, Australia, on December 15, 2014.
 \item \textbf{Charlie Hebdo shooting:} Two brothers forced their way into the offices of the French satirical weekly newspaper Charlie Hebdo in Paris, killing 11 people and wounding 11 more, on January 7, 2015.
 \item \textbf{Germanwings plane crash:} A passenger plane from Barcelona to D\"usseldorf crashed in the French Alps on March 24, 2015, killing all passengers and crew on board. The plane was ultimately found to have been deliberately crashed by the co-pilot of the plane.
\end{itemize}

In this case, we perform 5-fold cross-validation, having four events in the training set and the remaining event in the test set for each fold.

\begin{table}
\centering
\tbl{Counts of tweets with supporting, denying or questioning labels in each event collection on the PHEME dataset.\label{table:DatacountsPheme}}{
\begin{tabular}{lrrrrr}
\toprule 
Dataset & Rumours & Supporting & Denying & Questioning \\
\midrule 
Ottawa shooting & 58 & 161 & 76 & 63 \\
Ferguson riots& 46 & 192 & 82 & 94 \\
Charlie Hebdo & 74 & 235 & 56 & 51 \\
Germanwings crash & 68 & 67 & 12 & 28 \\
Sydney siege & 71 & 222 & 89 & 99 \\
\midrule
Total & 287 & 877 & 315 & 335 \\
\bottomrule 
\end{tabular}}
\end{table}

\section{Experiment Settings}
\label{sec:experiment-settings}

This section details the features and evaluation measures used in our experiments on tweet level stance classification.

\subsection{Classifiers}
\label{ssec:classifiers}

We begin by describing the classifiers we use for our experimentation, including Gaussian Processes, as well as a set of competitive baseline classifiers that we use for comparison.

\subsubsection{Gaussian Processes for Classification}
\label{sssec:gp}

Gaussian Processes are a Bayesian non-parametric machine learning framework that has been shown to work well for a range of NLP problems, often beating other state-of-the-art methods \cite{Cohn13modellingannotator,DBLP:conf/eacl/LamposAPC14,beck-etal_EMNLP:2014,jobs15acl}. 

A Gaussian Process defines a prior over functions, which combined with the likelihood of data points gives rise to a posterior over functions explaining the data. The key concept is a kernel function, which specifies how outputs correlate as a function of the input. Thus, from a practitioner's point of view, a key step is to choose an appropriate kernel function capturing the similarities between inputs.

We use Gaussian Processes as this probabilistic kernelised framework avoids the need for expensive cross-validation for hyperparameter selection.\footnote{There exist frequentist kernel methods, such as SVMs, which additionally require extensive heldout parameter tuning.} Instead, the marginal likelihood of the data can be used for hyperparameter selection.

The central concept of Gaussian Process Classification (GPC; \cite{Rasmussen:2005:GPM:1162254}) is a latent function $f$ over inputs \mbox{$\mathbf{x}$: $f(\mathbf{x}) \sim\ \mathcal{GP}(m(\mathbf{x}), k(\mathbf{x}, \mathbf{x}'))$}, where $m$ is the mean function, assumed to be $0$ and $k$ is the kernel function, specifying the degree to which the outputs covary as a function of the inputs. We use a linear kernel,
$k(\mathbf{x}, \mathbf{x}') = \sigma^2 \mathbf{x}^{\top}\mathbf{x}'$.
The latent function is then mapped by the probit function $\Phi(f)$ into the range $[0, 1]$, such that the resulting value can be interpreted as $p(y=1 | \mathbf{x})$.

The GPC posterior is calculated as

\begin{equation*}
p(f^* | X, \mathbf{y}, \mathbf{x_*}) = \int p(f^* | X, \mathbf{x_*}, \mathbf{f}) \frac{p(\mathbf{y} | \mathbf{f})p(\mathbf{f})}{p(\mathbf{y}|X)} d\mathbf{f} \, \!,
\end{equation*}
where  $p(\mathbf{y}|\mathbf{f}) = \displaystyle \prod_{j=1}^{n} \Phi(f_j)^{y_j} (1 - \Phi(f_j))^{1-y_j}$ is the Bernoulli likelihood of class $y$. After calculating the above posterior from the training data, this is used in prediction, i.e.,
\begin{equation*}
p(y_* \!=\! 1|X, \mathbf{y}, \mathbf{x_*}) \!=\!\!
\int \Phi\left(f_*\right)p\left(f_*|X, \mathbf{y}, \mathbf{x_*}\right)df_* \, .
\end{equation*}

The above integrals are intractable and approximation techniques are required to solve them. There exist various methods to deal with calculating the posterior; here we use Expectation Propagation (EP; \cite{Minka:2002:EGA:2073876.2073918}). In EP, the posterior is approximated by a fully factorised distribution, where each component is assumed to be an unnormalised Gaussian.

In order to conduct multi-class classification, we perform a one-vs-all classification for each label and then assign the one with the highest likelihood, amongst the three (supporting, denying, questioning). We choose this method due to interpretability of results, similar to recent work on occupational class classification \cite{jobs15acl}.


\paragraph*{Intrinsic Coregionalisation Model}
In the Leave-Part-Out (LPO) setting initial labelled tweets from the target rumour are observed as well, as opposed to the Leave-One-Out (LOO) setting.
In the case of LPO, we propose to weigh the importance of tweets from the reference rumours depending on how similar their characteristics are to the tweets from the target rumour available for training.
To handle this with GPC, we use a multiple output model based on the Intrinsic Coregionalisation Model (ICM; \cite{Alvarez:2012:KVF:2344402.2344403}). 
This model has already been applied successfully to NLP regression problems \cite{beck-etal_EMNLP:2014} and it can also be applied to classification ones. ICM parametrizes the kernel by a matrix which represents the extent of covariance between pairs of tasks. 
The complete kernel takes form of
\begin{equation*}
k((\mathbf{x}, d), (\mathbf{x}', d')) = k_{data}(\mathbf{x}, \mathbf{x}') B_{d, d'} \, ,
\end{equation*}

where B is a square coregionalisation matrix, $d$ and $d'$ denote the tasks of the two inputs and $k_{data}$ is a kernel for comparing inputs $\mathbf{x}$ and $\mathbf{x}'$ (here, linear). We parametrize the coregionalisation matrix \mbox{$B=\boldsymbol{\kappa} I+\boldsymbol{vv}^T$}, where $\boldsymbol{v}$ specifies the correlation between tasks and the vector $\mathbf{\kappa}$ controls the extent of task independence. 
Note that in case of LOO setting this model does not provide useful information, since no target rumour data is available to estimate similarity to other rumours.

\paragraph*{Hyperparameter selection}

We tune hyperparameters $\mathbf{v}$, $\boldsymbol{\kappa}$ and $\sigma^2$ by maximizing evidence of the model $p(\mathbf{y}|X)$, thus having no need for a validation set.

\paragraph*{Methods}
We consider GPs in three different settings, varying in what data the model is trained on and what kernel it uses. 
The first setting (denoted GP) considers only target rumour data for training. The second (GPPooled) additionally considers tweets from reference rumours (i.e. other than the target rumour). The third setting is GPICM, where an ICM kernel is used to weight influence from tweets from reference rumours.

\subsubsection{Baselines}
\label{sssec:baselines}

To assess and compare the efficiency of Gaussian Processes for rumour stance classification, we also experimented with five more baseline classifiers, all of which were implemented using the scikit Python package \cite{pedregosa2011scikit}: (1) \textit{majority classifier}, which is a naive classifier that labels all the instances in the test set with the most common class in the training set, (2) \textit{logistic regression} (MaxEnt), (3) \textit{support vector machines} (SVM), (4) \textit{naive bayes} (NB) and (5) \textit{random forest} (RF). The selection of these baselines is in line with the classifiers used in recent research on stance classification \cite{zeng2016unconfirmed}, who found that random forests, followed by logistic regression, performed best.

\subsection{Features}
\label{ssec:features}

We conducted a series of preprocessing steps in order to address data sparsity. All words were converted to lowercase; stopwords have been removed\footnote{We removed stopwords using the English list from Python's NLTK package.}; all emoticons were replaced by words\footnote{We used the dictionary from:
 \url{http://bit.ly/1rX1Hdk} and extended 
 it with: :o, $:|$, =/, :s, :S, :p.}; and stemming was performed. 
 In addition, multiple occurrences of a character were replaced with a double occurrence \cite{Agarwal:2011:SAT:2021109.2021114}, to correct for misspellings and lengthenings, e.g., \emph{looool}. 
 All punctuation was also removed, except for \emph{.}, \emph{!} and \emph{?}, which we hypothesize to be important for expressing emotion.
 Lastly, usernames were removed as they tend to be rumour-specific, i.e., very few users comment on more than one rumour.

After preprocessing the text data, we use either the resulting bag of words (BOW) feature representation and replace all words with their Brown cluster ids (Brown). Brown clustering is a hard hierarchical clustering method \cite{PercyLiangPhd}. It clusters words based on maximizing the probability of the words under the bigram language model, where words are generated based on their clusters.
In previous work it has been shown that Brown clusters yield better performance than directly using the BOW features \cite{Lukasik15_qazvinian}.
 
In our experiments, the clusters used were obtained using 1000 clusters acquired from a large scale Twitter corpus \cite{Owoputi13improvedpart-of-speech}, from which we can learn Brown clusters aimed at representing a generalisable Twitter vocabulary. Retweets are removed from the training set to prevent bias \cite{LLEWELLYN14.845}. More details on the Brown clusters that we used as well as the words that are part of each cluster are available online\footnote{http://www.cs.cmu.edu/~ark/TweetNLP/cluster\_viewer.html}.

During the experimentation process, we also tested additional features, including the use of the bag of words instead of the Brown clusters, as well as using word embeddings trained from the training sets \cite{mikolov2013efficient}. However, results turned out to be substantially poorer than those we obtained with the Brown clusters. We conjecture that this was due to the little data available to train the word embeddings; further exploring use of word embeddings trained from larger training datasets is left future work. In order to focus on our main objective of proving the effectiveness of a multi-task learning approach, as well as for clarity purposes, since the number of approaches to show in the figures increases if we also consider the BOW features, we only show results for the classifiers relying on Brown clusters as features.

\subsection{Evaluation Measures}
\label{ssec:eevaluation-measures}

Accuracy is often deemed a suitable evaluation measure to assess the performance of a classifier on a multi-class classification task. However, the classes are clearly imbalanced in our case, with varying tendencies towards one of the classes in each of the rumours. We argue that in these scenarios the sole evaluation based on accuracy is insufficient, and further measurement is needed to account for category imbalance. This is especially necessary in our case, as a classifier that always predicts the majority class in an imbalanced dataset will achieve high accuracy, even if the classifier is useless in practice. To tackle this, we use both micro-averaged and macro-averaged F1 scores. Note that the micro-averaged F1 score is equivalent to the well-known accuracy measure, while the macro-averaged F1 score complements it by measuring performance assigning the same weight to each category.

Both of the measures rely on precision (Equation \ref{eq:prec}) and recall (Equation \ref{eq:rec}) to compute the final F1 score.

\begin{equation}
 \text{Precision}_k = \frac{tp_k}{tp_k+fp_k}
 \label{eq:prec}
\end{equation}

\begin{equation}
 \text{Recall}_k = \frac{tp_k}{tp_k+fn_k}
 \label{eq:rec}
\end{equation}

where $tp_k$ (true positives) refer to the number of instances correctly classified in class $k$, $fp_k$ is the number of instances incorrectly classified in class $k$, and $fn_k$ is the number of instances that actually belong to class $k$ but were not classified as such.

The above equations can be used to compute precision and recall for a specific class. Precision and recall for all the classes in a problem with $c$ classes are computed differently if they are microaveraged (see Equations \ref{eq:prec-micro} and \ref{eq:rec-micro}) or macroaveraged (see Equations \ref{eq:prec-macro} and \ref{eq:rec-macro}).

\begin{equation}
 \text{Precision}_{\text{micro}} = \frac{\sum_{k = 1}^{c} tp_k}{\sum_{k = 1}^{c} tp_k + \sum_{k = 1}^{c} fp_k}
 \label{eq:prec-micro}
\end{equation}

\begin{equation}
 \text{Recall}_{\text{micro}} = \frac{\sum_{k = 1}^{c} tp_k}{\sum_{k = 1}^{c} tp_k + \sum_{k = 1}^{c} fn_k}
 \label{eq:rec-micro}
\end{equation}

\begin{equation}
 \text{Precision}_{\text{macro}} = \frac{\sum_{k = 1}^{c} \text{Precision}_k}{c}
 \label{eq:prec-macro}
\end{equation}

\begin{equation}
 \text{Recall}_{\text{macro}} = \frac{\sum_{k = 1}^{c} \text{Recall}_k}{c}
 \label{eq:rec-macro}
\end{equation}

After computing microaveraged and macroveraged precision and recall, the final F1 score is computed in the same way, i.e., calculating the harmonic mean of the precision and recall in question (see Equation \ref{eq:f1}).

\begin{equation}
 \text{F1} = \frac{2 \times \text{Precision} \times \text{Recall}}{\text{Precision} + \text{Recall}}
 \label{eq:f1}
\end{equation}

After computing the F1 score for each fold, we compute the micro-averaged score across folds.

\section{Results}
\label{sec:results}

First, we look at the results on each dataset separately. Then we complement the analysis by aggregating the results from both datasets, which leads to further understanding the performance of our classifiers on rumour stance classification.

\subsection{Comparison of Classifiers}
\label{ssec:comparison-classifiers}

We show the results for the LOO and LPO settings in the same figure, distinguished by the training size displayed in the X axis. In all the cases, labelled tweets from the remainder of the rumours (rumours other than the test/targer rumour) are used for training, and hence the training size shown in the X axis is in addition to those.
Note that the training size refers to the number of labelled instances that the classifier is making use of from the target rumour. Thus, a training size of 0 indicates the LOO setting, while training sizes from 10 to 50 pertain to the LPO setting. 

\begin{figure*}
  \centering
  \includegraphics[width=\textwidth]{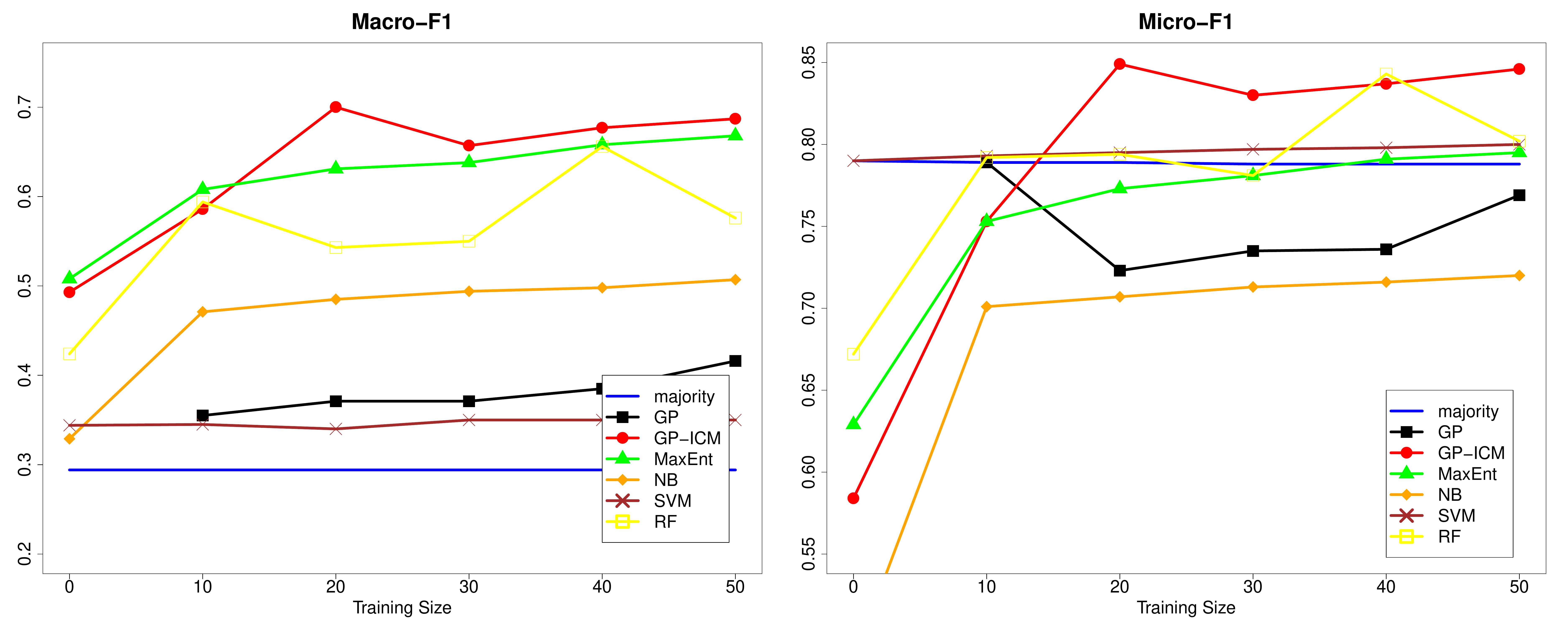}
  \caption{Micro-F1 and Macro-F1 scores for different methods versus the size of the target rumour used for training in the LPO setting on the England riots dataset. The test set is fixed to all but the first 50 tweets of the target rumour.}
  \label{fig:riots-f1}
\end{figure*}

Figure~\ref{fig:riots-f1} and Table~\ref{tab:riots-f1} show how micro-averaged and macro-averaged F1 scores for the England riots dataset change as the number of tweets from the target rumour used for training increases. We observe that, as initially expected, the performance of most of the methods improves as the number of labelled training instances from the target rumour increases. This increase is especially remarkable with the GP-ICM method, which gradually increases after having as few as 10 training instances. GP-ICM's performance keeps improving as the number of training instances approaches 50\footnote{Note that 50 tweets represent, on average, less than 7\% of the whole rumour, with the rest of the rumour yet to be observed.} Two aspects stand out from analysing GP-ICM's performance:

\begin{itemize}
 \item It performs poorly in terms of micro-averaged F1 when no labelled instances from the target rumour are used. However, it makes very effective use of the labelled training instances, overtaking the rest of the approaches and achieving the best results. This proves the ability of GP-ICM to make the most of the labelled instances from the target rumour, which the rest of the approaches struggle with.
 \item Irrespective of the number of labelled instances, GP-ICM is robust when evaluated in terms of macro-averaged F1. This means that GP-ICM is managing to determine the distribution of classes effectively, assigning labels to instances in the test set in a way that is better distributed than the rest of the classifier.
\end{itemize}

Despite the saliency of GP-ICM, we notice that two other baseline approaches, namely MaxEnt and RF, achieve competitive results that are above the rest of the baselines, but still perform worse than GP-ICM.

\begin{table}[hb]
\begin{center}
\tbl{Micro-F1 and Macro-F1 scores for different methods on the England riots dataset.\label{tab:riots-f1}}{
\begin{tabular}{ l | r r r r r r | r r r r r r }
\toprule
& \multicolumn{6}{c |}{Macro-F1} & \multicolumn{6}{c}{Micro-F1} \\
\midrule
& 0 & 10 & 20 & 30 & 40 & 50 & 0 & 10 & 20 & 30 & 40 & 50 \\
\midrule
Majority & 0.294 & 0.294 & 0.294 & 0.294 & 0.294 & 0.294 & \textbf{0.79} & \textbf{0.789} & 0.789 & 0.788 & 0.788 & 0.788 \\
GP &  & 0.355 & 0.371 & 0.371 & 0.385 & 0.416 &  & \textbf{0.789} & 0.723 & 0.735 & 0.736 & 0.769 \\
GP-ICM & 0.493 & 0.586 & \textbf{0.7} & \textbf{0.657} & \textbf{0.677} & \textbf{0.687} & 0.584 & 0.753 & \textbf{0.849} & \textbf{0.83} & 0.837 & \textbf{0.846} \\
MaxEnt & \textbf{0.508} & \textbf{0.608} & 0.631 & 0.638 & 0.658 & 0.668 & 0.629 & 0.753 & 0.773 & 0.781 & 0.791 & 0.795 \\
NB & 0.329 & 0.471 & 0.485 & 0.494 & 0.498 & 0.507 & 0.485 & 0.701 & 0.707 & 0.713 & 0.716 & 0.72 \\
SVM & 0.344 & 0.345 & 0.34 & 0.35 & 0.35 & 0.35 & 0.79 & 0.793 & 0.795 & 0.797 & 0.798 & 0.8 \\
RF & 0.424 & 0.594 & 0.543 & 0.55 & 0.656 & 0.576 & 0.672 & 0.792 & 0.794 & 0.781 & \textbf{0.843} & 0.802 \\
\bottomrule
\end{tabular}}
\end{center}
\end{table}

\begin{figure*}
  \centering
  \includegraphics[width=\textwidth]{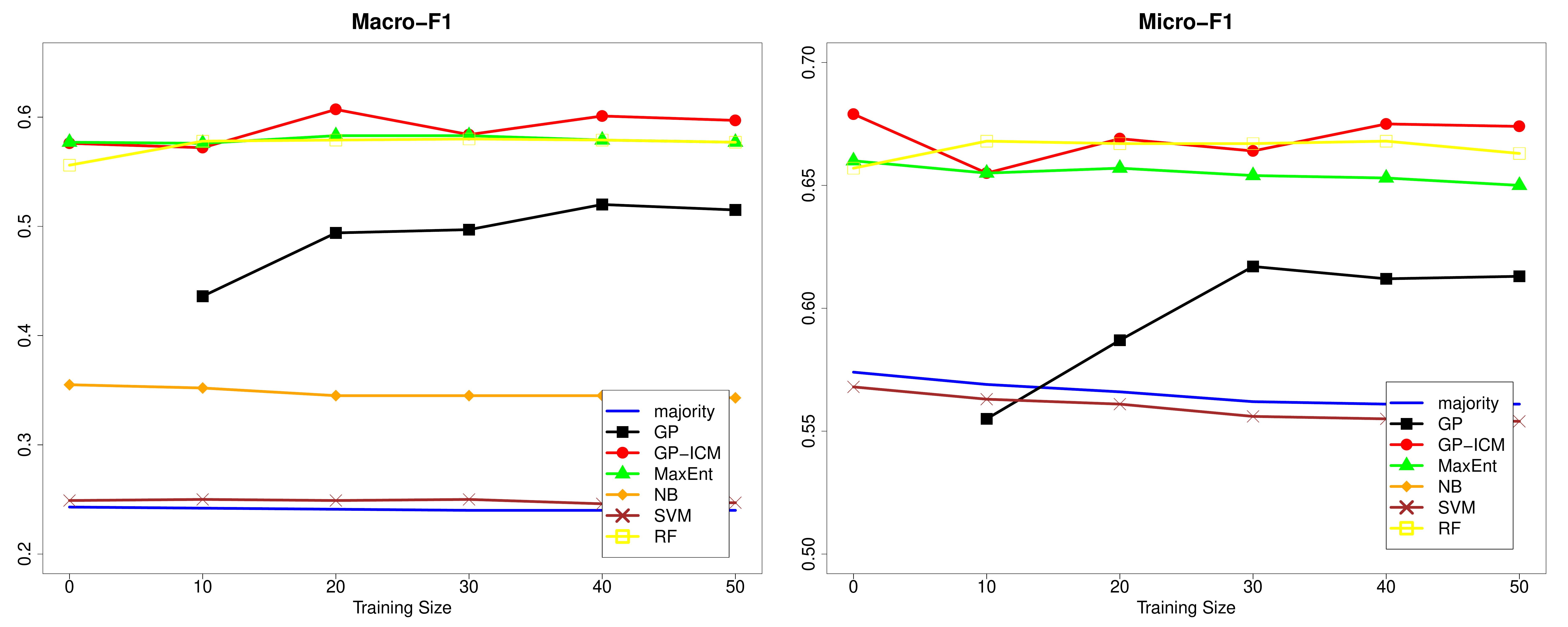}
  \caption{Micro-F1 and Macro-F1 scores for different methods versus the size of the target rumour used for training in the LPO setting on the PHEME dataset. The test set is fixed to all but the first 50 tweets of the target rumour.}
  \label{fig:pheme-f1}
\end{figure*}

The results from the PHEME dataset are shown in  Figure~\ref{fig:pheme-f1} and Table~\ref{tab:pheme-f1}. Overall, we can observe that results are lower in this case than they were for the riots dataset. The reason for this can be attributed to the following two observations: on the one hand, each fold pertaining to a different event in the PHEME dataset means that the classifier encounters a new event in the classification, where it will likely find new vocabulary, which may be more difficult to classify; on the other hand, the PHEME dataset is more prominently composed of tweets that are replying to others, which are likely shorter and less descriptive on their own and hence more difficult to get meaningful features from. Despite the additional difficulty in this dataset, we are interested in exploring if the same trend holds across classifiers, from which we can generalise the analysis to different types of classifiers.

One striking difference with respect to the results from the riots dataset is that, in this case, the classifiers, including GP-ICM, are not gaining as much from the inclusion of labelled instances from the target rumour. This is likely due to the heterogeneity of each of the events in the PHEME dataset. Here a diverse set of rumourous newsworthy pieces of information are discussed pertaining to the selected events as they unfold. By contrast, each rumour in the riots dataset is more homogeneous, as each rumour focuses on a specific story.

Interestingly, when we compare the performance of different classifiers, we observe that GP-ICM again outperforms the rest of the approaches, both in terms of micro-averaged and macro-averaged F1 scores. While the micro-averaged F1 score does not increase as the number of training instances increases, we can see a slight improvement in terms of macro-averaged F1. This improvement suggests that GP-ICM does still take advantage of the labelled training instances to boost performance, in this case by better distributing the predicted labels.

Again, as we observed in the case of the riots dataset, two baselines stand out, MaxEnt and RF. They are very close to the performance of GP-ICM for the PHEME dataset, event outperforming it in a few occasions. 
In the following subsection we take a closer look at the differences among the three classifiers.

\begin{table}[hb]
\begin{center}
\tbl{Micro-F1 and Macro-F1 scores for different methods on the PHEME dataset.\label{tab:pheme-f1}}{
\begin{tabular}{ l | r r r r r r | r r r r r r }
\toprule
& \multicolumn{6}{c |}{Macro-F1} & \multicolumn{6}{c}{Micro-F1} \\
\midrule
& 0 & 10 & 20 & 30 & 40 & 50 & 0 & 10 & 20 & 30 & 40 & 50 \\
\midrule
Majority & 0.243 & 0.242 & 0.241 & 0.24 & 0.24 & 0.24 & 0.574 & 0.569 & 0.566 & 0.562 & 0.561 & 0.561 \\
GP &  & 0.436 & 0.494 & 0.497 & 0.52 & 0.515 &  & 0.555 & 0.587 & 0.617 & 0.612 & 0.613 \\
GP-ICM & 0.576 & 0.572 & \textbf{0.607} & \textbf{0.584} & \textbf{0.601} & \textbf{0.597} & \textbf{0.679} & 0.655 & \textbf{0.669} & 0.664 & \textbf{0.675} & \textbf{0.674} \\
MaxEnt & \textbf{0.577} & \textbf{0.576} & 0.583 & 0.583 & 0.579 & 0.577 & 0.66 & 0.655 & 0.657 & 0.654 & 0.653 & 0.65 \\
NB & 0.355 & 0.352 & 0.345 & 0.345 & 0.345 & 0.343 & 0.371 & 0.373 & 0.366 & 0.363 & 0.363 & 0.358 \\
SVM & 0.249 & 0.25 & 0.249 & 0.25 & 0.246 & 0.247 & 0.568 & 0.563 & 0.561 & 0.556 & 0.555 & 0.554 \\
RF & 0.556 & 0.578 & 0.579 & 0.58 & 0.579 & 0.577 & 0.657 & \textbf{0.668} & 0.667 & \textbf{0.667} & 0.668 & 0.663 \\
\bottomrule
\end{tabular}}
\end{center}
\end{table}

\subsection{Analysing the Performance of the Best-Performing Classifiers}
\label{ssec:delving-best-classifiers}

We delve into the results of the best-performing classifiers, namely GP-ICM, MaxEnt and RF, looking at their per-class performance. This will help us understand when they perform well and where it is that GP-ICM stands out achieving the best results.

Tables \ref{tab:per-class-riots} and \ref{tab:per-class-pheme} show per-class F1 measures for the aforementioned three best-performing classifiers for the England riots dataset and the PHEME dataset, respectively. They also show statistics of the mis-classifications that the classifiers made, in the form of percentage of deviations towards the other classes.

Looking at the per-class performance analysis, we observe that the performance of GP-ICM varies when we look into Precision and Recall. Still, in all the dataset-class pairs, GP-ICM performs best in terms of either Precision or Recall, even though never in both. Moreover, it is generally the best in terms of F1, achieving the best Precision and Recall. The only exception is with MaxEnt classifying questioning tweets more accurately in terms of F1 for the England riots.

When we look at the deviations, we see that all the classifiers suffer from the datasets being imbalanced towards supporting tweets. This results in all classifiers classifying numerous instances as supporting, while they are actually denying or questioning. This is a known problem in rumour diffusion, as previous studies have found that people barely deny or question rumours but generally tend to support them irrespective of their actual veracity value \cite{zubiaga2016analysing}. While we have found that GP-ICM can tackle the imbalance issue quite effectively and better than other classifiers, this caveat posits the need for further research in dealing with the striking majority of supporting tweets in the context of rumours in social media.

\begin{table}[ht]
 \begin{center}
  \tbl{Per-class precision, recall and F1 scores for the best-performing classifiers on the England riots dataset.\label{tab:per-class-riots}}{
  \begin{tabular}{ l l | r r r | r r r }
   \toprule
   Class & Classifier & \multicolumn{3}{ c |}{Performance} & \multicolumn{3}{ c }{Deviations} \\
   \midrule
   & & P & R & F1 & S & D & Q \\
   \midrule
   \multirow{3}{*}{supporting (S)} & GP-ICM & 0.887 & \textbf{0.931} & \textbf{0.909} & --- & 0.86\% & 6.03\% \\
   & MaxEnt & \textbf{0.926} & 0.818 & 0.869 & --- & 11.60\% & 6.56\% \\
   & RF & 0.859 & 0.914 & 0.885 & --- & 1.21\% & 7.41\% \\
   \midrule
   \multirow{3}{*}{denying (D)} & GP-ICM & \textbf{0.862} & 0.483 & \textbf{0.619} & 49.52\% & --- & 2.22\% \\
   & MaxEnt & 0.500 & \textbf{0.739} & 0.597 & 22.69\% & --- & 3.41\% \\
   & RF & 0.696 & 0.311 & 0.430 & 67.51\% & --- & 1.43\% \\
   \midrule
   \multirow{3}{*}{questioning (Q)} & GP-ICM & \textbf{0.478} & 0.608 & 0.535 & 34.28\% & 4.92\% & --- \\
   & MaxEnt & 0.461 & \textbf{0.647} & \textbf{0.538} & 29.48\% & 5.78\% & --- \\
   & RF & 0.367 & 0.472 & 0.413 & 42.00\% & 10.79\% & --- \\
   \bottomrule
  \end{tabular}}
 \end{center}
\end{table}

\begin{table}[ht]
 \begin{center}
  \tbl{Per-class precision, recall and F1 scores for the best-performing classifiers on the PHEME dataset.\label{tab:per-class-pheme}}{
  \begin{tabular}{ l l | r r r | r r r }
   \toprule
   Class & Classifier & \multicolumn{3}{ c |}{Performance} & \multicolumn{3}{ c }{Deviations} \\
   \midrule
   & & P & R & F1 & S & D & Q \\
   \midrule
   \multirow{3}{*}{supporting (S)} & GP-ICM & \textbf{0.731} & 0.825 & \textbf{0.776} & --- & 7.12\% & 10.34\% \\
   & MaxEnt & 0.715 & 0.814 & 0.761 & --- & 12.08\% & 6.55\% \\
   & RF & 0.696 & \textbf{0.860} & 0.770 & --- & 7.42\% & 6.55\% \\
   \midrule
   \multirow{3}{*}{denying (D)} & GP-ICM & \textbf{0.540} & 0.313 & \textbf{0.396} & 50.36\% & --- & 18.35\% \\
   & MaxEnt & 0.427 & \textbf{0.325} & 0.369 & 50.54\% & --- & 16.97\% \\
   & RF & 0.494 & 0.285 & 0.362 & 58.48\% & --- & 13.00\% \\
   \midrule
   \multirow{3}{*}{questioning (Q)} & GP-ICM & 0.594 & \textbf{0.647} & \textbf{0.619} & 27.21\% & 8.13\% & --- \\
   & MaxEnt & 0.635 & 0.569 & 0.600 & 29.54\% & 13.52\% & --- \\
   & RF & \textbf{0.657} & 0.552 & 0.600 & 34.16\% & 10.68\% & --- \\
   \bottomrule
  \end{tabular}}
 \end{center}
\end{table}

\section{Discussion}

Experimentation with two different approaches based on Gaussian Processes (GP and GP-ICM) and comparison with respect to a set of competitive baselines over two rumour datasets enables us to gain generalisable insight on rumour stance classification on Twitter. This is reinforced by the fact that the two datasets are very different from each other. The first dataset, collected during the England riots in 2011, is a single event that we have split into folds, each fold belonging to a separate rumour within the event; hence, all the rumours are part of the same event. The second dataset, collected within the PHEME project, includes tweets for a set of five newsworthy events, where each event has been assigned a separate fold; therefore, the classifier needs to learn from four events and test on a new, unknown event, which has proven more challenging.

Results are generally consistent across datasets, which enables us to generalise conclusions well. We observe that while GP itself does not suffice to achieve competitive results, GP-ICM does instead help boost the performance of the classifier substantially to even outperform the rest of the baselines in the majority of the cases.

GP-ICM has proven to consistently perform well in both datasets, despite their very different characteristics, being competitive not only in terms of micro-averaged F1, but also in terms of macro-averaged F1. GP-ICM manages to balance the varying class distributions effectively, showing that its performance is above the rest of the baselines in accurately determining the distribution of classes. This is very important in this task of rumour stance classification, owing to the fact that even if a classifier that is 100\% accurate is unlikely, a classifier that accurately guesses the overall distribution of classes can be of great help. If a classifier makes a good estimation of the number of denials in an aggregated set of tweets, it can be useful to flag those potentially false rumours with high level of confidence.

Another factor that stands out from GP-ICM is its capacity to perform well when a few labelled instances of the target rumour are leveraged in the training phase. GP-ICM effectively exploits the knowledge garnered from the few instances from the target rumour, outperforming the rest of the baselines even when its performance was modest when no labelled instances were used from the target rumour.

In light of these results, we deem GP-ICM the most competitive approach to use when one can afford to get a few instances labelled from the target rumour. 
The labels from the target rumour can be obtained in practice in different ways: (1) having someone in-house (e.g. journalists monitoring breaking news stories) label a few instances prior to running the classifier, (2) making use of resources for human computation such as crowdsourcing platforms to outsource the labelling work, or (3) developing techniques that will attempt to classify the first few instances, incorporating in the training set those for which a classification with high level of confidence has been produced. The latter presents an ambitious avenue for future work that could help alleviate the labelling task.

On the other hand, in the absence of labelled data from the target rumour, which is the case of the LOO setting, the effectiveness of the GP-ICM classifier is not as prominent. For this scenario, other classifiers such as MaxEnt and Random Forests have proven more competitive and one could see them as better options. However, we do believe that the remarkable difference that the reliance on the LPO setting produces is worth exploiting where possible.

\section{Conclusions}

Social media is becoming an increasingly important tool for maintaining social resilience: individuals use it to express opinions and follow events as they unfold; news media organisations use it as a source to inform their coverage of these events; and government agencies, such as the emergency services, use it to gather intelligence to help in decision-making and in advising the public about how they should respond \cite{procter2013readingb}. While previous research has suggested that mechanisms for exposing false rumours are implicit in the ways in which people use social media \cite{procter2013readinga}, it is nevertheless critically important to explore if there are ways in which computational tools can help to accelerate these mechanisms so that misinformation and disinformation can be targeted more rapidly, and the benefits of social media to society maintained \cite{derczynski2015pheme}.

As a first step to achieving this aim, this \work has investigated the problem of classifying the different types of stance expressed by individuals in tweets about rumours. First, we considered a setting where no training data from the target rumours is available (LOO). Without access to annotated examples of the target rumour the learning problem becomes very difficult. We showed that in the supervised domain adaptation setting (LPO), even annotating a small number of tweets helps to achieve better results. Moreover, we demonstrated the benefits of a multi-task learning approach, as well as that Brown cluster features are more useful for the task than simple bag of words.

Findings from previous work, such as \citeauthor{Mendoza-credibility-2013,procter2013readinga}, have suggested that the aggregate stance of individual users is correlated with actual rumour veracity. Hence, the next step in our own work will be to make use of the classifier for the stance expressed in the reactions of individual Twitter users in order to predict the actual veracity of the rumour in question. Another interesting direction for future work would be the addition of non-textual features to the classifier. For example, the rumour diffusion patterns \cite{lukasik15} may be a useful cue for stance classification.

\begin{acks}
This work is partially supported by the European Union under grant agreement No. 611233 {\sc Pheme}. The work was implemented using the GPy toolkit \cite{gpy2014}. This research utilised Queen Mary's MidPlus computational facilities, supported by QMUL Research-IT and funded by EPSRC grant EP/K000128/1.
\end{acks}

\bibliographystyle{ACM-Reference-Format-Journals}
\bibliography{bibliography}
\end{document}